\documentclass[11pt]{article}
\pdftrailerid{}
\usepackage[final]{acl}
\usepackage{times}
\usepackage{latexsym}
\usepackage[T1]{fontenc}
\usepackage[utf8]{inputenc}
\usepackage{microtype}
\usepackage{inconsolata}
\usepackage{graphicx}
\usepackage{amsmath}
\usepackage{amssymb}
\usepackage{booktabs}
\usepackage{multirow}
\usepackage{array}
\usepackage{xcolor}
\usepackage{float}
\usepackage{placeins}
\usepackage{marvosym}

\newcolumntype{L}[1]{>{\raggedright\arraybackslash}p{#1}}
\newcommand{\runin}[1]{%
  \par\smallskip\noindent\mbox{\textbf{#1:}}\enspace\ignorespaces}

\makeatletter
\let\acllegacylnmakecol\@LN@makecol
\RemoveFromHook{build/column/before}[lineno]
\AtBeginDocument{\let\@LN@makecol\acllegacylnmakecol}
\makeatother

\title{Does the Selected Object Reach the Reader? Auditing Identity Handoffs in Grounded Language-Model Pipelines}
\author{
  \textbf{Siddharth Vohra\textsuperscript{1,2}}\thanks{Work done by Siddharth
  Vohra does not relate to the position he currently holds at Amazon Web
  Services AI Native.\quad $^{\dagger}$Project lead.\quad
  \textsuperscript{\Letter}Corresponding author.},
  \textbf{Runmin Jiang\textsuperscript{1,$\dagger$}},
  \textbf{Xiaomo Li\textsuperscript{1}},
  \textbf{Min Xu\textsuperscript{1,\Letter}}
\\
\\
  \textsuperscript{1}Carnegie Mellon University, Pittsburgh, PA, USA \\
  \textsuperscript{2}Amazon Web Services AI Native, Pittsburgh, PA, USA
\\
  \small{
    \texttt{\{svohra, runminj, xiaomol, mxu1\}@andrew.cmu.edu}
  }
}

\begin{document}
\maketitle

\begin{abstract}
Grounded language-model pipelines can be divided into three stages: selecting
an object, retrieving passages for it, and using that evidence to answer. If the
selected object must reach the reader, losing it breaks the handoff. Benchmark
recall checks the dataset-linked object, which can differ. We audit 600 HybridQA
questions across three selector families. On 1,463 resolvable
records where the selected object matches the dataset-traced passage, exact key
lookup and exact title matching return the object every time. With every ranked
rule given the same decoded selected title, body-only BM25 omits it on 389
records (26.6\%) at cutoff five, while hybrid retrieval with reranking omits it
on 14 (1.0\%). The two identities differ on 329 of 1,792
resolvable records. With original-question rankings, their top-five checks
disagree on 106 records (5.9\%). Frozen reader comparisons associate the aligned
object's presence with 28.6 to 31.0 points higher exact match. In a deliberately
selected 64-item cohort, removing that passage sharply lowers exact match, while
removing a similar-length comparison passage does not reproduce the drop. We
release the Returned-Object Profile (ROP), an executable record of the target,
returned-ID field, cutoff, membership rule, and complete expected population,
with data and an offline replay.
\end{abstract}

\section{Introduction}
\label{sec:intro}

Some grounded language-model pipelines turn a structured choice into text
evidence. An entity linker can select a page, and a table reasoner can choose a
linked record. Retrieval then turns that choice into passages for a reader
\citep{li2020elq,chen2020hybridqa,shavarani2025entity}. The reader receives only
the returned passages. If the selected object drops from that list, the rest of
the pipeline can no longer use it.

A pipeline can fail because selection is poor, because a required identity
handoff breaks, or because the reader fails to use returned evidence. These
causes call for different fixes. A single top-$k$ recall score cannot separate
them unless it names both the object being checked and the component responsible
for returning it.

We call the selected object $U$ and the passage linked by the dataset $G$.
Dataset-trace recall checks benchmark coverage with $G$. Selected-object return
checks whether a selector-to-retriever handoff delivers $U$. Semantic
correctness remains its own measurement. When $U=G$, both checks ask about the
same object. When the identities differ, the score must say whether it describes
benchmark coverage or delivery of the recorded selection.

\begin{figure*}[t]
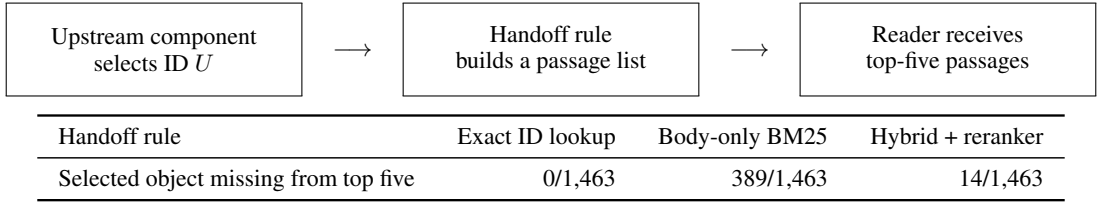

\centering
{\small
\begin{tabular}{c@{\hspace{4mm}$\longrightarrow$\hspace{4mm}}c@{\hspace{4mm}$\longrightarrow$\hspace{4mm}}c}
\fbox{\parbox[c][10mm][c]{0.23\textwidth}{\centering Upstream component\\selects ID $U$}}
& \fbox{\parbox[c][10mm][c]{0.23\textwidth}{\centering Handoff rule\\builds a passage list}}
& \fbox{\parbox[c][10mm][c]{0.23\textwidth}{\centering Reader receives\\top-five passages}}
\end{tabular}}
\par\vspace{2mm}
{\footnotesize
\setlength{\tabcolsep}{8pt}
\begin{tabular}{lrrr}
\toprule
Handoff rule & Exact ID lookup & Body-only BM25 & Hybrid + reranker \\
\midrule
Selected object missing from top five & 0/1,463 & 389/1,463 & 14/1,463 \\
\bottomrule
\end{tabular}}
\caption{For each of 1,463 aligned, resolvable records, the selected object is
held constant across handoff rules. The rule determines whether that object
appears in the reader's top five passages. Both ranked rules receive the same
decoded selected title.}
\label{fig:handoff-overview}
\end{figure*}

Our main question is what happens after a pipeline already holds a selected
object ID. It can look up that ID directly or use its title or text to retrieve
a broader passage set. These handoff rules preserve the recorded identity at
different rates. Figure~\ref{fig:handoff-overview} shows the interface and the
main result.

We evaluate this handoff on 600 HybridQA questions. Three model families each
process all 600 questions, giving 1,800 planned model-question records. On 1,463
records, the selected object agrees with HybridQA's linked
passage and resolves in the frozen corpus. Exact key dereference and exact
decoded-title equality return all 1,463 objects. At cutoff five,
body-only BM25 omits 389, ColBERTv2 omits 20, and the final hybrid reranker
omits 14. With the selected ID held fixed, the handoff rule determines whether
that ID appears in the evidence list.

The 329 resolvable records where $U$ and $G$ differ show why this boundary needs
its own target. They come from 160 questions. Original-question BM25 receives
only the question, yet the two targets give different top-five labels on 106 of
1,792 records. Queries that name the selected object produce more disagreements,
showing why query construction belongs in the evaluation record. This audit
measures which identity reaches the reader. Semantic quality remains a separate
measurement.

We also follow aligned objects into the reader. In recorded five-passage
comparisons, exact match is 28.6 to 31.0 points higher when the shared object is
present. The contexts differ in other ways, so this result is associative. We
then make fresh calls for 64 high-contrast cases chosen before the new outcomes.
Exact match falls when the aligned passage is removed. Removing a similar-length
nontraced passage does not reproduce the loss.

We make three contributions. First, we compare nine ways to turn a selected ID
into reader evidence and measure whether the object reaches the top five.
Second, we score selected-object return and dataset-trace recall against the
same frozen rankings across four query forms, five retrieval stacks, and three
cutoffs. Third, we release the ROP schema and tested reference evaluator, which
encode the target, expected population, identity resolver, output field, cutoff,
membership rule, and run result. The released artifact reproduces every reported
aggregate offline from the frozen rankings and reader records. The project
website is available at \url{https://opensciagent.github.io/rop}.

\section{Measuring an Identity Handoff}
\label{sec:framework}

Some grounding pipelines carry a selected object ID into retrieval. The
upstream component emits that ID, and retrieval returns a ranked list of passage
or chunk IDs. This gives the boundary a concrete input and output. We map both
through one identity resolver. Selected-object return
at cutoff $k$ succeeds when one of the first $k$ returned IDs maps to $U$ or to
a declared acceptable alternative. Dataset-trace recall uses the same ranking
and membership rule with $G$ as the target.

The Returned-Object Profile (ROP) records this evaluation. It begins with the
interface claim. When a documented policy requires the selected object to
remain available after retrieval, ROP turns that rule into a complete-run check.
The same fields can describe an interface with no selected-return requirement.
In either mode, ROP isolates one boundary: whether the declared target appears
in the returned IDs.

For a developer who requires the selected object in the reader context, these
counts distinguish identity-preserving lookup from ranked evidence construction
that needs an explicit return check.

\begin{table}[htbp]
\centering
\footnotesize
\setlength{\tabcolsep}{2.2pt}
\caption{The five decisions recorded by a ROP declaration.
Appendix~\ref{app:formal} gives the complete fields and verdict order.}
\label{tab:contract}
\begin{tabular}{L{2.25cm}L{4.45cm}}
\toprule
Evaluation question & Decision recorded in the declaration \\
\midrule
Which claim applies? & The documented interface and whether it requires selected-object return. \\
Which object counts? & The selected target or its declared acceptable alternatives. \\
Which records belong? & The complete list of expected record IDs. \\
What counts as returned? & The identity map, returned-ID field, cutoff, and \texttt{any} or \texttt{all} membership rule. \\
How is the result decided? & One verdict for every expected record and one status for the complete run. \\
\bottomrule
\end{tabular}
\end{table}

A resolver maps each raw key, alias, or chunk ID to one shared object identity.
The returned-ID field and cutoff identify the part of the output being checked.
The \texttt{any} rule accepts one returned ID that resolves to an acceptable
target. The \texttt{all} rule requires every raw resolver member for one
acceptable target. Unknown IDs do not match, and duplicate IDs still consume
ranked positions.

The expected population is equally important. Suppose a run expects two rows
and both pass. If one row disappears, the pass fraction over the surviving row
is still 1.0. The run is only half complete. ROP lists every expected record ID,
assigns one verdict to each expected record, and marks missing, duplicate,
malformed, or unexpected observations before it reports the run result.

\begin{table}[t]
\centering
\scriptsize
\setlength{\tabcolsep}{1.8pt}
\caption{Population accounting in the synthetic hard-mode fixture. The
observed-row pass fraction is identical, while the run status detects the
missing expected record.}
\label{tab:population-example}
\begin{tabular}{lrrrrl}
\toprule
Case & Expected & Observed & Pass & Pass fraction & Run status \\
\midrule
Complete & 2 & 2 & 2 & 1.0 & \texttt{RUN\_PASS} \\
One missing & 2 & 1 & 1 & 1.0 & \texttt{RUN\_UNVERIFIABLE} \\
\bottomrule
\end{tabular}
\end{table}

Table~\ref{tab:population-example} shows the missing-row case. The reference
suite also separates a failed membership check from an audit that cannot be
scored. A resolvable target outside the cutoff receives \texttt{FAIL}. A missing
or malformed returned-ID list receives \texttt{UNVERIFIABLE}. Missing expected
records make the run unverifiable, while duplicate or unexpected record IDs
invalidate it. These outcomes point to different parts of the pipeline.

\begin{table}[t]
\centering
\scriptsize
\setlength{\tabcolsep}{2.2pt}
\renewcommand{\arraystretch}{1.12}
\caption{How the reference evaluator accounts for common observations. Record
accounting and run status remain separate so that an omitted target, unusable
output, and incomplete run cannot collapse into one score.}
\label{tab:fixture-coverage}
\begin{tabular}{L{2.8cm}L{1.8cm}L{2.3cm}}
\toprule
Observed condition & Record accounting & Run status \\
\midrule
Requirement inactive & \texttt{NOT\_APPLICABLE} & \texttt{NOT\_APPLICABLE} \\
Target returned in a complete run & \texttt{PASS} & \texttt{RUN\_PASS} \\
Target absent & \texttt{FAIL} & \texttt{RUN\_FAIL} \\
Returned IDs missing or malformed & \texttt{UNVERIFIABLE} & \texttt{RUN\_UNVERIFIABLE} \\
Expected record missing & Missing population entry & \texttt{RUN\_UNVERIFIABLE} \\
Duplicate or unexpected record ID & Invalid population & \texttt{RUN\_INVALID} \\
\bottomrule
\end{tabular}
\end{table}

A membership failure means that a usable returned list omits the target. An
unverifiable record lacks a usable returned list. A missing expected ID makes
the run incomplete, while duplicate or unexpected IDs make its observed
population inconsistent with the declaration. Keeping these outcomes separate
tells a developer whether to inspect retrieval, logging, or run assembly.

The released synthetic fixtures exercise all seven record verdicts and five run
statuses. They establish the behavior of the reference evaluator. The next
section reports a retrospective audit over 1,800 frozen HybridQA records using
the same target, resolver, cutoff, and population concepts. The artifact replays
that audit offline from hash-bound rankings and reader records.
Appendix~\ref{app:formal} gives the schema and verdict order.
Appendix~\ref{app:repro} gives the execution settings and replay steps.

\section{HybridQA Identity-Handoff Audit}
\label{sec:evaluation}

\subsection{Study design}

HybridQA supports an identity-handoff audit because its table cells link to
Wikipedia passages \citep{chen2020hybridqa}. We retain questions with one
answer-containing linked passage, no answer in the table, no linked-cell label
in the question, and a selection rule fixed before model outputs are inspected.
These filters retain 7,201 training questions. A salted hash selects 600, with
at most one question per table. The corpus contains 286,270 passages linked
from 15,314 tables.

Claude Opus 4.7, GPT-5.5, and Gemini 3.1 Pro Preview each receive a question and
its full table. Each model is asked to return a selected \texttt{/wiki/...} key
$U$ and a free-form retrieval fact $F$. The selected key is the upstream
identity in the frozen pipeline. Each query condition determines whether
retrieval receives that identity. HybridQA's annotated link $G$ remains a
separate trace object.
Each model-question pair contributes one recorded selection. The three model
families describe three frozen pipelines, so this design does not measure how a
selector varies across repeated samples.
Six of the 1,800 selector rows are invalid, and two valid keys do not resolve in
the corpus. All eight remain in the accounting.

Table~\ref{tab:population-accounting} accounts for every planned selector row
before retrieval is scored.

\begin{table}[t]
\centering
\scriptsize
\setlength{\tabcolsep}{2.1pt}
\caption{Complete accounting for the 1,800 planned model-question records. The
four outcome columns partition every row. The $U=G$ and $U\ne G$ columns contain
corpus-resolvable selected IDs.}
\label{tab:population-accounting}
\begin{tabular}{@{}lrrrrr@{}}
\toprule
Model & Planned & \shortstack{Invalid\\output} & \shortstack{ID absent\\from corpus} & $U=G$ & $U\ne G$ \\
\midrule
Claude & 600 & 0 & 1 & 456 & 143 \\
GPT & 600 & 6 & 0 & 509 & 85 \\
Gemini & 600 & 0 & 1 & 498 & 101 \\
\midrule
Total & 1,800 & 6 & 2 & 1,463 & 329 \\
\bottomrule
\end{tabular}
\end{table}

We compare nine ways to turn a selected ID into evidence. Two deterministic
baselines use exact key lookup or exact decoded-title equality. The seven ranked
rules include three BM25 variants over titles, bodies, or both, plus \mbox{BGE-M3},
\mbox{ColBERTv2}, reciprocal-rank fusion, and hybrid reranking
\citep{robertson2009bm25,chen2024bgem3,santhanam2022colbertv2,bge2024reranker}.
Exact key lookup bypasses ranking and maps $U$ directly to its passage. Exact
title equality decodes the selected key and uses a verified one-to-one map from
corpus titles back to object IDs. The ranked rules use that same decoded title
as a query, then let every corpus object compete for the returned positions.
These two anchors separate identity resolution from the ranking step.

Dense retrieval uses \mbox{BGE-M3}. \mbox{ColBERTv2} compares query and passage
tokens. Reciprocal-rank fusion combines both models with BM25 at $k=60$
\citep{cormack2009rrf}. The final stack reranks the top 100 fused candidates.
We request up to 20 returned IDs and score the recorded lists at cutoffs 1, 5,
and 20.

All selector outputs and rankings were frozen before this audit. We score $U$
and $G$ against exactly the same returned IDs, so the target is the only part of
the comparison that changes. Appendix~\ref{app:repro} gives the selector prompt,
execution settings, pinned model revisions, retriever settings, and full
accounting.

\subsection{The handoff rule determines object return}

We hold the target fixed. On 1,463 records, the selected ID is resolvable
and equals the dataset trace, so every rule is checked against the same object.
All seven ranked rules receive the same one-to-one decoded selected title. This
comparison isolates delivery of an explicit identity. Original-question
retrieval is evaluated separately.

\begin{table}[htbp]
\centering
\footnotesize
\setlength{\tabcolsep}{2.6pt}
\caption{Selected-object absence for two deterministic baselines and seven
ranked retrieval rules. Every record has one fixed, resolvable target.}
\label{tab:executor-contract}
\begin{tabular}{L{3.25cm}rr}
\toprule
Handoff rule & Missing at 5 & Missing rate \\
\midrule
Exact key lookup & 0/1,463 & .000 \\
Exact title equality & 0/1,463 & .000 \\
Title-only BM25 & 66/1,463 & .045 \\
Title + body BM25 & 284/1,463 & .194 \\
Body-only BM25 & 389/1,463 & .266 \\
BGE-M3 & 65/1,463 & .044 \\
ColBERTv2 & 20/1,463 & .014 \\
Hybrid RRF & 148/1,463 & .101 \\
Hybrid + reranker & 14/1,463 & .010 \\
\bottomrule
\end{tabular}
\end{table}

\begin{table}[htbp]
\centering
\scriptsize
\setlength{\tabcolsep}{1.0pt}
\caption{Cutoff-five misses by model family on the same aligned records used in
Table~\ref{tab:executor-contract}. The target is fixed within every row.}
\label{tab:delivery-by-model}
\begin{tabular}{lrrrrrr}
\toprule
Model & $n$ & BM25 & BGE & ColB. & RRF & Final \\
\midrule
Claude & 456 & 125 & 20 & 6 & 47 & 4 \\
GPT & 509 & 133 & 23 & 7 & 51 & 5 \\
Gemini & 498 & 131 & 22 & 7 & 50 & 5 \\
\midrule
All & 1,463 & 389 & 65 & 20 & 148 & 14 \\
\bottomrule
\end{tabular}
\end{table}

The deterministic rows verify that the resolver and title projection preserve
identity on all 1,463 records. The seven ranked rules omit between 14 and 389 at
cutoff five. Body-only BM25 loses the target on 26.6\% of records, compared with
1.4\% for ColBERTv2 and 1.0\% for the final reranker. The selected ID and
evaluation target stay fixed throughout. The handoff rule determines whether
that ID appears in the reader's evidence list.

The pattern repeats across all three model families. Across the 1,463 aligned
paired records, the final reranker retains 1,071 body-only BM25 hits and
recovers 378 BM25 misses. It loses three BM25 hits, and both rules miss on 11
records. The comparison therefore isolates changes made by the evidence builder
after selection.

The final reranker misses 14 selected objects across five questions. Six appear
at rank seven or eight, beyond the reader's cutoff. Eight are absent from both
released top-20 lists. Under a documented selected-return requirement, six
misses occur near the cutoff and are consistent with a cutoff adjustment. The
remaining eight require deeper retrieval or explicit retention of the selected
object to certify its return.

Appendix~\ref{app:sensitivity} reports the corpus-wide title sanity check,
remaining cutoffs, and per-model results.

\subsection{Reading the final reranker misses}

The final reranker leaves 14 misses in the aligned cohort. Every row in this
analysis has a resolvable selected ID and $U=G$, so the selected-object and
dataset-trace checks agree on the target. The misses come from five questions
and form two clear groups in the released rankings.

\begin{table}[t]
\centering
\footnotesize
\setlength{\tabcolsep}{2.5pt}
\caption{The 14 aligned records missed by the final reranker at cutoff five.
Ranks are identical across the listed model rows. $>20$ means absent from the
released top 20.}
\label{tab:final-miss-cases}
\begin{tabular}{L{3.55cm}rrr}
\toprule
Selected object & Rows & RRF & Final \\
\midrule
Hypo-Arena & 3 & $>20$ & $>20$ \\
Philadelphia Athletics & 3 & 5 & 7 \\
Defender (association football) & 3 & 8 & 8 \\
Apple Vale, Queensland & 3 & $>20$ & $>20$ \\
SS Sylvan Arrow & 2 & $>20$ & $>20$ \\
\bottomrule
\end{tabular}
\end{table}

Six misses sit just below the reader cutoff. Philadelphia Athletics appears at
rank five after fusion and moves to rank seven after reranking for all three
model records. Defender appears at rank eight before and after reranking. These
objects remain in the recorded list, so changing the cutoff would change their
handoff result.

The other eight misses are absent from both released top-20 lists. This group
contains all three Hypo-Arena records, all three Apple Vale records, and two SS
Sylvan Arrow records. The stored depth does not separate a rank below 20 from a
complete lexical or neural nonmatch. ROP records the observable endpoint result
without assigning an unseen cause.

\subsection{Why the evaluation target matters}

Across 1,792 resolvable model-question records, the selected and dataset-traced
identities differ on 329 records drawn from 160 of the 600 questions. We score
each frozen ranking twice, once against $U$ and once against $G$. The ranking,
cutoff, and identity resolver do not change.
The same 600 questions appear once for each model family. Pooled model-question
counts describe the recorded audit and are not independent question samples.

The four body-BM25 queries expose the selected ID at different points. The
original question is shared across selector families, and $U$ is not an input.
The direct rewrite is generated from the question and full table, again without
receiving $U$ as the selected ID. The retrieval fact is co-generated with $U$
and is prompted to name the same entity. Its executed BM25 terms match the
decoded-title terms on 291 of the 329 object-distinct records. The decoded-title
query is built directly from $U$ using the fixed title projection. The last two
forms are controlled selected-name delivery tests.

\begin{table}[t]
\centering
\scriptsize
\setlength{\tabcolsep}{1.4pt}
\caption{Target-aware cutoff-five BM25 results on the 329 records where $U$
and $G$ differ. Different is the sum of the two one-object columns.}
\label{tab:lossy-query-joint}
\begin{tabular}{@{}lrrrrr@{}}
\toprule
Query & Both & \shortstack{$U$\\only} & \shortstack{$G$\\only} & Neither & Different \\
\midrule
Original question & 19 & 78 & 28 & 204 & 106 \\
Direct rewrite & 48 & 140 & 35 & 106 & 175 \\
Retrieval fact & 43 & 213 & 4 & 69 & 217 \\
Decoded title & 42 & 210 & 3 & 74 & 213 \\
\bottomrule
\end{tabular}
\end{table}

Original-question BM25 receives only the question, yet its two membership checks
differ on 106 records, or 5.9\% of the resolvable cohort. It returns neither
object on 204 of the 329 object-distinct records. A direct rewrite gives 175
different labels. The retrieval-fact and decoded-title conditions deliberately
name $U$ and give 217 and 213. Their larger counts show how query construction
changes the magnitude and balance of target disagreement.

The selected-name stress test compares five stacks under the same decoded-title
query. Selected-object return at cutoff five ranges from .725 to .992 across
model families and stacks. Under the two targets, complete-cohort stack order
remains unchanged for every model family and cutoff in the reported sensitivity
checks. The target changes record-level diagnosis in this study, while aggregate
stack choice stays stable. Appendix~\ref{app:sensitivity} reports the full
cutoffs, per-model breakdowns, and an exact ranking example.

\FloatBarrier
\section{From Returned Evidence to Answers}
\label{sec:reader}

An object that reaches the evidence list can now influence the reader. We study
this step on records where the selected and dataset-traced IDs agree, leaving
one shared passage to follow from retrieval into the answer.

\subsection{Passage presence and answer accuracy}

The frozen answer matrix has three reader prompts with no answer hint. Their
evidence comes from retrieval over the decoded selected ID, a model-written
rewrite, or the original question. For each pair, we keep records where $U=G$,
both prompts receive five passages, and the shared object appears in one passage
list but not the other. Exact match uses standard HybridQA normalization.

\begin{table}[t]
\centering
\scriptsize
\setlength{\tabcolsep}{2.2pt}
\caption{Hint-free answer associations when $U=G$. Return EM scores the context
that contains the shared object, and Omit EM scores the paired context that does
not. Intervals resample question IDs and keep all model-family rows for each
sampled question together.}
\label{tab:delivery-switch-answer}
\begin{tabular}{lrrrr}
\toprule
Arm pair & $n$ & Return EM & Omit EM & $\Delta$EM [95\%] \\
\midrule
ID/question & 832 & 79.2 & 50.6 & 28.6 [24.5,32.8] \\
ID/rewrite & 200 & 76.0 & 45.0 & 31.0 [23.5,38.9] \\
Rewrite/question & 903 & 81.9 & 52.6 & 29.3 [25.4,33.4] \\
\bottomrule
\end{tabular}
\end{table}

Across all three pairs, exact match is 28.6 to 31.0 points higher in the context
that contains the shared object. The comparison remains associative because the
paired contexts also change passage identities, ranks, lengths, distractors,
and input length. We therefore isolate one passage in a deletion experiment.

\subsection{Deleting the aligned passage}

The deletion study uses 64 $U=G$ items selected from earlier outputs. Each
historical item was exact with the shared passage and inexact without it in both
the retrieved-context and passage-only comparisons. We fixed these cases before
the fresh calls, creating a high-contrast cohort for testing whether the same
passage-specific pattern repeats.

Each item receives three intact repetitions and three trace-drop repetitions.
The intact arm keeps the recorded request content. Trace-drop removes the
aligned passage and leaves all other prompt content in place. For 59 items,
sham-drop instead removes the nontraced passage closest in rendered length that
does not contain the normalized reference answer. We measure the change in exact
match between intact and trace-drop. Appendix~\ref{app:repro} gives the task,
retry, and failure accounting.

\begin{table}[t]
\centering
\footnotesize
\setlength{\tabcolsep}{2.6pt}
\caption{Passage-deletion contrasts within the preselected 64-item
high-contrast cohort. I, T, and S denote intact, trace-drop, and sham-drop.
Values are three-repeat item means with paired bootstrap intervals.}
\label{tab:passage-deletion}
\begin{tabular}{lrrr}
\toprule
Contrast & $n$ & EM A/B & $\Delta$EM [95\%] \\
\midrule
I/T & 64 & 92.7/15.1 & 77.6 [68.8,85.4] \\
S/T & 59 & 96.0/14.7 & 81.4 [71.8,89.8] \\
I/S & 59 & 92.1/96.0 & $-4.0$ [$-9.0$,1.1] \\
\bottomrule
\end{tabular}
\end{table}

Within this fixed cohort, removing the aligned passage lowers exact match by
77.6 points, and sham-drop exceeds trace-drop by 81.4 points. The intact versus
sham interval includes zero. All 24 terminal failures remain in their planned
cells and score zero. Even under the most adverse reassignment of those
failures, both trace-removal gaps remain at least 73.4 points. Majority exact is
63 of 64 for intact, 57 of 59 for sham-drop, and 9 of 64 for trace-drop.
Answer-token F1 gives corresponding gaps of 48.5 and 52.1 points.

\section{Related Work}
\label{sec:related}

Grounding research studies which evidence a language model retrieves and how
that evidence supports its output. RAG supplies retrieved context for generation
\citep{lewis2020rag}. RAGChecker separates retrieval and generation errors, and
MIRAGE links answer tokens to retrieved documents
\citep{ru2024ragchecker,qi2024mirage}. CF-RAG uses counterfactual queries to
distinguish causal evidence from correlated distractors, while AgenticRAGTracer
diagnoses multi-step retrieval reasoning
\citep{qin2026cfrag,you2026agenticragtracer}.
Rationale studies test whether explanations reflect model behavior, while
causal mediation measures how components affect a chosen outcome
\citep{deyoung2020eraser,madsen2024self,vig2020mediation}. These methods inspect
evidence support and answer behavior. Our audit follows a named object across
one component boundary before the reader sees it.

Retrieval inputs also change across a pipeline. ReAct, IRCoT, STEP, and
Iter-RetGen use intermediate reasoning or generated text to guide later search
\citep{yao2023react,trivedi2023ircot,luo2022step,shao2023iterative}. Query
rewriting, HyDE, and RaFe learn or generate new retrieval inputs
\citep{ma2023query,gao2023hyde,mao2024rafe}. Our four query forms show that query
construction must be reported alongside the profile because it determines how
directly retrieval is exposed to the selected identity.

Identity-bearing grounding systems expose this handoff because they pass named
entities or passages across stages. Entity linking resolves mentions to entity
identities \citep{wu2020scalable,li2020elq}. Graph and entity retrieval then use
entity or passage identities to collect evidence for later components
\citep{min2019knowledge,sun2019pullnet,shavarani2025entity}.
A unified data-to-text interface converts structured knowledge into retrievable
text \citep{ma2022unified}. HybridQA supplies tables with linked passages, and
multi-row table-text QA draws evidence from both modalities
\citep{chen2020hybridqa,kumar2023multirow}. OTT-QA retrieves tables and text,
while TableRAG retrieves heterogeneous text and table chunks
\citep{chen2021ottqa,yu2025tablerag}. ROP applies at a boundary only when the
interface exposes one of these identities and documents that it must remain in
the returned set.

Provenance, tracing, and software contracts provide general ways to record and
validate pipeline state. PROV represents provenance information
\citep{groth2013prov}. \mbox{OpenTelemetry}'s generative-AI conventions define
retrieval fields \citep{opentelemetry2026traces}. Design by Contract and OCL
express assertions and behavioral constraints
\citep{meyer1992contract,omg2014ocl}. Typed data-validation systems encode type,
presence, value-count, and domain constraints in schemas
\citep{breck2019validation}. ROP specializes these general mechanisms for a
retrieval boundary by declaring the selected target, identity resolver,
returned-ID field, cutoff, expected population, and verdict order. The
reference evaluator scores every expected record and the complete run.

\section{Discussion}

The study begins with a simple interface fact. A reader can only use the
passages it receives. Once an upstream component emits an object ID, the
pipeline exposes a concrete boundary: $U$ enters retrieval, and ranked IDs leave
it. Table~\ref{tab:executor-contract} shows that the rule between those two
endpoints matters even when the query states the selected identity directly.
Selected-object return measures this boundary and identifies where a recorded
choice disappears.

The aligned and object-distinct analyses play different roles. On the 1,463
aligned records, $U$ and $G$ identify the same passage. The omission counts in
Table~\ref{tab:executor-contract} therefore measure the handoff rule without any
target disagreement. This is the main implementation result. The 329
object-distinct records answer a reporting question: two valid membership tests
can describe different components of the same ranking. The reader analysis
returns to aligned records and follows one shared passage into the answer. These
three pieces form one chain from selected ID to returned evidence to reader
behavior.

A complete pipeline evaluation can place a score at each point in that chain.
Selector evaluation records whether $U$ fits the task. A selected-return miss
identifies an endpoint omission after selection. A wrong answer after both
checks pass leaves evidence quality and reader behavior to inspect. This
division gives each score one component and one repair target.

Dataset-trace recall answers a second useful question. It asks whether retrieval
covers the object linked by the benchmark. Selected-object return asks whether
retrieval delivers the upstream choice. The 106 different labels under the
original-question query show that these questions remain distinct even when
retrieval never sees $U$. Reporting the target makes clear whether a score
describes benchmark coverage or selector-to-retriever delivery.

Direct lookup and ranked retrieval serve different purposes. Lookup preserves
the selected identity, while ranking adds neighboring passages or chunks. A
pipeline with a documented selected-return requirement can reserve one reader
slot for $U$ and fill the remaining slots from ranked retrieval. ROP checks the
reserved identity. Relevance and answer quality remain separate evaluations.
The exact-lookup rows establish identity resolution, while the ranked rows
measure what evidence construction retains or displaces.

Query construction must be reported alongside the profile because it determines
how directly retrieval is exposed to the selected identity. The decoded-title
test exposes $U$ directly, while the original question and direct rewrite do
not receive it.
The target disagreement changes from 106 records with the original question to
213 with the decoded title. That difference is part of the pipeline behavior.
A reported handoff result therefore needs the target, query construction,
returned-ID field, and cutoff together.

Complete population accounting closes a second gap. A pass fraction over rows
that survived processing cannot reveal a missing row. The two-row example in
Table~\ref{tab:population-example} keeps the surviving pass fraction at 1.0,
while ROP marks the incomplete run as unverifiable. This matters whenever API
failures, parser errors, or interrupted jobs can remove planned records before
aggregation.

ROP hard mode is tied to a documented selected-return requirement. For other
interfaces, the same fields provide a descriptive audit without assigning a
hard pass or failure. The synthetic fixture tests the evaluator and all verdict
paths. The \mbox{HybridQA} study supplies the separate empirical stress test. Together,
they show how to state the requirement, count the full run, and measure whether
the selected identity reaches the reader.

\FloatBarrier
\section{Conclusion}

An identity-bearing pipeline can hold a resolvable object ID and still lose it
when building evidence for the reader. On 1,463 HybridQA records where the
selected object matches the dataset trace, the two deterministic handoff rules
return every object. The seven ranked rules omit between 14 and 389 at cutoff
five. Measuring this boundary shows whether the recorded choice survives the
rule that turns it into passages.

Dataset-trace recall measures benchmark coverage. Selected-object return
measures whether the recorded choice survives retrieval. On identical
original-question BM25 rankings, the two checks give different top-five labels
on 106 of 1,792 resolvable records, or 5.9\%. Query construction changes that
disagreement rate. Together, the two targets identify which component the score
describes.

ROP turns a documented selected-return requirement into a complete-run test. The
reference evaluator binds the target, returned field, cutoff, resolver, and
expected population, assigns a verdict to every record, and assigns a status to
the complete run. The HybridQA audit identifies which handoff rules omit the
selected object. ROP makes the chosen rule and denominator explicit and
replayable.

\section*{Limitations}

ROP v1 checks endpoint membership in supplied, hash-bound records. It verifies
byte consistency and agreement between observed IDs and the declared expected
population. It does not establish when the declaration was created or who
produced the logs. Those checks require external controls. The released
hard-mode example uses a synthetic interface.

The empirical study covers one English Wikipedia table-text corpus and uses one
recorded selector outcome per model-question pair, so it does not estimate
selector stability. The 329 exact-ID mismatches have no semantic annotations.
The deletion result applies to its deliberately selected 64-item cohort.
Selector quality, whether both identities provide valid evidence, and average
answer benefit require separate evaluations.

\section*{Ethical Considerations}

The study uses the public, Wikipedia-derived \mbox{HybridQA} dataset and collects no new
personal data. We recruited no human annotators. We do not treat the generated
answers as verified facts or evaluate them for social bias. The release retains
invalid outputs and item-level traces for inspection.

\section*{Acknowledgments}

This work was supported in part by U.S. NIH grant R35GM158094.

\bibliography{references}

\appendix

\section{Audit Records and Replay}
\label{app:formal}

\subsection{How ROP Evaluates a Run}

ROP evaluates one handoff. An upstream component supplies a selected key or
declared alternatives, and a downstream component supplies ranked raw corpus
IDs. Each run begins with a declaration, the instruction sheet that fixes what
the evaluator will check. In hard mode, the declaration also lists every record
ID expected in the run. The evaluator resolves the identities, tests endpoint
membership, assigns one verdict to each observed record, accounts for missing
expected IDs, and assigns one run status.

\runin{Set the mode and population}
The declaration names the target interface, interface version, policy authority,
producer, and policy revision. Hard mode applies exactly when the documented
interface specification requires the selected object to remain available at the
downstream endpoint. A valid soft or open declaration returns
\texttt{NOT\_APPLICABLE}. Any accompanying measurements remain separate
diagnostics. Prompt text preserves the prompt used for the run. The declaration
alone determines the mode.

ROP v1 uses the authority and producer named in the declaration. SHA-256 hashes
verify that the evaluated files contain the same bytes as the supplied files.
They do not authenticate the producer. Independent verification of event order
and timing requires signed timestamps. Signed timestamps are not part of v1.

In hard mode, the expected population is the complete list of record IDs
required for the run. The declaration stores this population as a sorted list,
its count, identifier field, and SHA-256 hash. Its metadata states that the list
was fixed before returned IDs were produced. ROP verifies the supplied bytes and
internal consistency. Timing is established outside the evaluator. The
evaluator compares every observed row with that list and retains every row.
Missing IDs appear in
\texttt{missing\_record\_ids} and in the \texttt{MISSING} histogram. The
evaluator also detects duplicate IDs, malformed IDs, and unexpected records.
A diagnostic pass fraction uses the evaluable records. The run status uses the
declared population and accounts for every expected ID.

The JSON Schema validates manifest syntax. The evaluator then validates
relationships across files and fields. The declaration envelope is the
top-level wrapper around the instruction sheet. These checks cover the
envelope, every SHA-256 hash, interface and component equality, mode and policy
revision, population order and count, resolver ownership, whether each witness
decodes to a distinct selection, and every referenced file. The schema and
these semantic checks together determine the evaluation result.

\runin{Resolve identities and test the ranking}
A versioned resolver $D_v$ maps each raw key, alias, or chunk ID to one
canonical object. Let $\mathcal U_i$ contain the selected key or its declared
alternatives. The evaluator reads the first $k$ raw IDs returned by the
component named in the declaration, where
$\widehat R_i@k=(r_1,\ldots,r_k)$. Every value in $\mathcal U_i$ must resolve,
giving $S_i=\{D_v(u):u\in\mathcal U_i\}$. ROP applies the following rules to
each canonical object $o$:
\begin{align}
q_{\mathrm{any}}(o,\widehat R@k)
  &= \mathbf 1[\exists r\in\widehat R@k:D_v(r)=o],\\
q_{\mathrm{all}}(o,\widehat R@k)
  &= \mathbf 1[M_v(o)\subseteq \operatorname{set}(\widehat R@k)].
\end{align}
Here $M_v(o)$ is the resolver's nonempty set of raw members for object $o$. The
evaluator applies the declared rule to one acceptable object at a time. The
membership test succeeds exactly when $q(o,\widehat R@k)=1$ for at least one
$o\in S_i$. Under \texttt{any}, aliases, object IDs, and member IDs first map to
their canonical object, and one match is sufficient. Under \texttt{all}, every
raw resolver member for one acceptable object must appear. The evaluator
applies the cutoff before either rule. An alias or member assigned to two
canonical owners invalidates the resolver. Unknown returned IDs are nonmatches.
Duplicate IDs keep their ranked positions and consume cutoff slots.

In v1, the string fields \texttt{corpus\_uri} and
\texttt{composite\_corpus\_key} are passed to the resolver without
interpretation. The evaluator only applies their declared nonempty-string
constraint.

\runin{Decode a bound witness}
An evaluation record carries identity in one of two forms. It either stores the
raw selected key or stores a declared invertible witness with a hash-bound exact
inverse map. A bound lookup table decodes the witness back to the selected key.
In witness mode, each record carries a SHA-256 hash over the binding profile,
declaration ID and exact manifest hash, population file hash and record ID,
witness value, decoded selection, and inverse-map file hash. The lookup table
must be one-to-one: no two witness values decode to the same selection. An
invalid or non-injective lookup table, or a decoded selection outside the
resolver, produces \texttt{DECLARATION\_INVALID}. An unknown witness value
produces \texttt{RESOLUTION\_ERROR}. A missing, malformed, or incorrectly
hash-bound record witness produces \texttt{SELECTOR\_ERROR}. These rules
establish endpoint membership from either the carried selected key or the
selection decoded from the bound witness. The hashes verify the supplied bytes
and do not authenticate the producer. A lineage extension that records each
intermediate handoff and its verified parent remains future work. The endpoint
check returns either a membership label or a specific identity, resolution, or
returned-log error. The complete ROP evaluator validates the declaration and
files, accounts for every expected ID, assigns record verdicts, and produces
the run status.

\runin{Assign one verdict to each record}
The evaluator applies the ordered checks in
Table~\ref{tab:record-verdicts}. Some records receive a terminal verdict before
the membership test. Each observed record receives exactly one of seven
terminal verdicts:
\texttt{PASS}, \texttt{FAIL}, \texttt{SELECTOR\_ERROR},
\texttt{RESOLUTION\_ERROR}, \texttt{UNVERIFIABLE},
\texttt{DECLARATION\_INVALID}, or \texttt{NOT\_APPLICABLE}.
The first matching condition assigns the final verdict, and no later check
overwrites it. In hard mode, every expected ID is represented by an observed
record or by an explicit \texttt{MISSING} entry in the run histogram and
missing-ID list.

\begin{table}[t]
\centering
\footnotesize
\setlength{\tabcolsep}{2.2pt}
\begin{tabular}{L{3.1cm}L{3.7cm}}
\toprule
Record verdict & First condition that applies \\
\midrule
\texttt{DECLARATION\_INVALID} & Top-level declaration envelope, instruction sheet, authority, policy, population, resolver, identity witness, output binding, or any referenced file fails validation \\
\texttt{NOT\_APPLICABLE} & Valid declaration selects soft or open mode \\
\texttt{SELECTOR\_ERROR} & Selected key or bound witness is missing, malformed, or incorrectly bound \\
\texttt{RESOLUTION\_ERROR} & Selected identity is absent from the verified resolver \\
\texttt{UNVERIFIABLE} & Returned-ID field is missing or malformed \\
\texttt{PASS} & Declared endpoint-membership rule is satisfied \\
\texttt{FAIL} & Declared endpoint-membership rule is violated \\
\bottomrule
\end{tabular}
\caption{Record verdicts in fixed order. The first condition that applies
assigns the final verdict.}
\label{tab:record-verdicts}
\end{table}

\runin{Assign one status to the run}
After assigning record verdicts, the evaluator gives the complete run one of
five statuses. It first validates the declaration. A valid soft or open
declaration yields \texttt{NOT\_APPLICABLE}. In hard mode, the evaluator
compares the observed IDs with the complete expected list and then combines the
expected record verdicts. Incomplete evidence receives its own run status.

\begin{table}[t]
\centering
\footnotesize
\setlength{\tabcolsep}{2.2pt}
\begin{tabular}{L{2.65cm}L{4.15cm}}
\toprule
Run status & First condition that applies \\
\midrule
\texttt{RUN\_INVALID} & Declaration invalid, or a hard-mode run contains malformed, duplicate, or unexpected observed IDs \\
\texttt{NOT\_APPLICABLE} & Valid declaration selects soft or open mode \\
\texttt{RUN\_FAIL} & Hard mode, no invalid-run condition, and at least one expected row has verdict \texttt{FAIL} \\
\texttt{RUN\_UNVERIFIABLE} & Hard mode, no invalid-run condition, and no \texttt{FAIL}. An expected ID is missing or an expected row has selector error, resolution error, or a missing or malformed returned-ID list \\
\texttt{RUN\_PASS} & Hard mode, observed population equals the sealed population, and every expected record passes \\
\bottomrule
\end{tabular}
\caption{Run statuses follow a fixed order. An invalid declaration yields
\texttt{RUN\_INVALID}. A valid soft or open declaration yields
\texttt{NOT\_APPLICABLE}. A hard run is then classified by population validity,
recorded failure, incomplete evidence, or complete success.}
\label{tab:run-verdicts}
\end{table}

\runin{Requirements for \texttt{RUN\_PASS}}
For a hard run, \texttt{RUN\_PASS} requires every check to succeed. Every
referenced file must match its SHA-256 hash. The declaration, instruction sheet,
policy, population, resolver, and any witness must pass their semantic checks.
The observed IDs must exactly equal the declared expected population. Each
expected ID appears once, with no malformed, duplicate, or unexpected
ID. Every expected record must satisfy the declared return condition. The run
status remains separate from the diagnostic pass fraction.

\runin{Fields retained for replay}
The released ledgers join on question identifier and, where applicable, model
family. Each bound row retains the selected identifier $U_i$, dataset-traced
identifier $G_i$, and raw retrieved document IDs $\widehat R$ for every query
and retriever. The stored ranking depth covers every reported cutoff. The
versioned resolver records corpus-resolution status and maps the raw IDs to the
object ranking $R$. Replay derives resolved identity agreement $C$,
selected-object output membership $P$, and trace-object output membership $H$.
The API-call ledgers retain terminal-attempt status and explicit invalid-output
fields. The artifact manifest and replay extractors bind each exact source
member by byte count and SHA-256 hash. The answer ledgers retain the ordered
evidence links and normalized answer scores used by the matched-five audit.

\runin{Validate joined records before analysis}
Before a joined row enters an analysis, the replay tool validates its bound
source rows and reconstructs every field used by that analysis. All rows must
share the same question and, where applicable, the same model. Replay also
checks the named condition, reference path, cutoff, logical task, terminal
attempt, invalid-output policy, source SHA-256 hashes, and status. A mismatch
rejects the joined row and leaves the declared denominator unchanged. A
schema-invalid terminal response or unissued dependency cell remains explicit
under the recorded denominator policy. The same released bytes and
deterministic extractors reproduce the joined rows and aggregate counts.

\runin{Measure overlap from joined rows}
Separate field totals do not identify their overlap. The released rows compute
every reported conjunction directly on a shared question-model key. The
analysis counts complete joined records. This study binds one dataset-traced
path per item. The executable selected-return profile declares set-valued
selection rules before scoring and accepts alias- or chunk-level returned IDs.
An evaluation with several acceptable benchmark traces must declare the trace
set before scoring and retain every within-path conjunction.

\FloatBarrier
\section{Reproducibility Details}
\label{app:repro}

This appendix records cohort construction, retrieval outputs, model settings,
and the open-weight check.

\subsection{Cohort construction}

We build the cohort from fixed rules that do not use model outputs. We normalize
each question by lowercasing its text, keeping matches of
\texttt{[a-z0-9]+}, and joining them with one space. We apply this
case-insensitive term list: \emph{most, least, highest, lowest, largest,
smallest, biggest, oldest, youngest, first, second, third, last, latest,
earliest, more than, less than, finished, winner, won, gold, silver, bronze,
position}, and \emph{ranked}. A salted SHA-256 hash of the question identifier
orders every item that passes the filters. We take the first 600, allowing at
most one question from any table.

The released files make this selection reproducible. They record the salt, the
source hash, each item's selection hash, and the final cohort hash. They also
freeze the IDs and hashes for all 7,201 training candidates and 455 development
candidates. The split denominators come directly from the pinned upstream
HybridQA files.

\runin{Keeping protocol development separate}
We reserved a hash-locked set of 120 questions for protocol development and
excluded those IDs from every later development cohort. These later cohorts
provide a separate in-dataset check. Before collecting the corresponding
model outcomes, we fixed the cohorts, prompts, answer conditions, retrieval
cutoffs, scoring rules, and treatment of invalid responses. The prompt also
fixed what \texttt{selected\_link} means.

\runin{Frozen selector instruction}
The system instruction states:
\begin{quote}\small
You select a bridge entity from a table. The bridge entity is the one
table-linked Wikipedia page whose passage is needed to answer the question.
Do not answer the question. Use the table and the question to choose exactly
one link that is explicitly shown. Return only the required JSON object.
\end{quote}
The question and full table appear before this final user instruction:
\begin{quote}\small
Choose the single shown link that best bridges the table to the external
information needed for the answer. Set \texttt{selected\_link} to the exact
\texttt{/wiki/...} value shown. Set \texttt{retrieval\_fact} to a concise
human-readable name for that same entity, suitable as a search query. Do not
include the final answer.
\end{quote}
This instruction defines $U$ for the retrospective comparison.

\runin{Synthetic hard-mode fixture}
The released hard fixture uses \texttt{fixture-ranker-api/v1}. It reads the
selected target from \texttt{selector.selected\_key}, reads returned IDs from
\texttt{object\_id}, checks the first three under \texttt{any}, and expects
exactly \texttt{run\_alpha} and \texttt{run\_beta}. The prompt is stored as
provenance and cannot activate or override hard mode.

Every analysis that compares the selected key with the dataset trace reuses the
same frozen identifiers, frozen ranking records, and answer records. The Qwen
analysis and the checks that use object identifiers directly have their own
hash-bound inputs. For each analysis, the release connects the exact inputs to
the code, execution metadata, and outputs that produced the reported result.

\subsection{Retrieval results across models and cutoffs}
\label{app:sensitivity}

The decoded-title query is a controlled selected-name test. It exposes $U$
through the verified one-to-one title projection and asks whether each stack
returns that identity.

Across the complete corpus inventory, exact title equality returns every object
at all three cutoffs. Title-only BM25 omits 15,379 objects at cutoff one, 7,667
at cutoff five, and 5,258 at cutoff 20. Of the 286,270 title queries, 141 become
empty after lexical preprocessing. This is a corpus-wide identity-resolution
sanity check without a selector.

\begin{table}[htbp]
\centering
\footnotesize
\setlength{\tabcolsep}{3.0pt}
\caption{Corpus-wide omission counts when every object is queried by its own
decoded title. Each column has denominator 286,270.}
\label{tab:corpus-roundtrip}
\begin{tabular}{lrrr}
\toprule
Handoff rule & At 1 & At 5 & At 20 \\
\midrule
Exact title equality & 0 & 0 & 0 \\
Title-only BM25 & 15,379 & 7,667 & 5,258 \\
\bottomrule
\end{tabular}
\end{table}

\begin{table}[htbp]
\centering
\footnotesize
\setlength{\tabcolsep}{2.2pt}
\caption{Selected-object return in the decoded-title test at three cutoffs.
Denominators are 599 for Claude, 594 for GPT, and 599 for Gemini.}
\label{tab:selected-cutoffs}
\begin{tabular}{@{}llrrr@{}}
\toprule
Retriever & Model & R@1 & R@5 & R@20 \\
\midrule
\multirow{3}{*}{Body BM25} & Claude & .506 & .725 & .840 \\
& GPT & .515 & .746 & .845 \\
& Gemini & .524 & .750 & .855 \\
\addlinespace
\multirow{3}{*}{BGE-M3} & Claude & .778 & .955 & .983 \\
& GPT & .773 & .958 & .988 \\
& Gemini & .775 & .955 & .988 \\
\addlinespace
\multirow{3}{*}{ColBERTv2} & Claude & .881 & .983 & .990 \\
& GPT & .896 & .988 & .992 \\
& Gemini & .891 & .987 & .992 \\
\addlinespace
\multirow{3}{*}{Hybrid RRF} & Claude & .674 & .891 & .967 \\
& GPT & .695 & .897 & .966 \\
& Gemini & .699 & .902 & .972 \\
\addlinespace
\multirow{3}{*}{RRF + rerank} & Claude & .855 & .990 & .995 \\
& GPT & .855 & .992 & .995 \\
& Gemini & .856 & .990 & .995 \\
\bottomrule
\end{tabular}
\end{table}

One frozen record shows how the target check works. The selector chooses Xavier
University as $U$, while HybridQA links Xavier Musketeers football as $G$. With
the shared query \texttt{Xavier University}, the two frozen handoff rules in
Table~\ref{tab:joined-example} give opposite top-five labels.

\begin{table}[htbp]
\centering
\footnotesize
\setlength{\tabcolsep}{2.5pt}
\begin{tabular}{lrrc}
\toprule
Handoff rule & Rank of $U$ & Rank of $G$ & Top-five result \\
\midrule
Body-only BM25 & 6 & 3 & $G$ only \\
BGE-M3 & 1 & 16 & $U$ only \\
\bottomrule
\end{tabular}
\caption{Frozen rankings for one Claude record. $U$ is Xavier University and
$G$ is Xavier Musketeers football.}
\label{tab:joined-example}
\end{table}

\begin{table}[htbp]
\centering
\footnotesize
\setlength{\tabcolsep}{1.5pt}
\caption{Cutoff-five results for the 329 object-distinct records under the
decoded-title query. Different labels is the sum of the two one-object columns.}
\label{tab:proxy-disagreement}
\begin{tabular}{@{}lrrrrr@{}}
\toprule
Stack & Both & \shortstack{$U$\\only} & \shortstack{$G$\\only} & Neither & \shortstack{Different\\labels} \\
\midrule
Body BM25 & 42 & 210 & 3 & 74 & 213 \\
BGE-M3 & 31 & 284 & 0 & 14 & 284 \\
ColBERTv2 & 39 & 285 & 0 & 5 & 285 \\
Hybrid RRF & 34 & 258 & 0 & 37 & 258 \\
RRF + rerank & 48 & 278 & 0 & 3 & 278 \\
\bottomrule
\end{tabular}
\end{table}

\begin{table*}[t]
\centering
\footnotesize
\setlength{\tabcolsep}{4.2pt}
\begin{tabular}{lrcccc}
\toprule
Stack & $k$ & Claude & GPT & Gemini & Pooled \\
\midrule
Body-only BM25 & 1 & 0/83/2/58 & 0/54/2/29 & 0/67/1/33 & 0/204/5/120 \\
 & 5 & 15/88/2/38 & 13/54/0/18 & 14/68/1/18 & 42/210/3/74 \\
 & 20 & 22/95/0/26 & 20/50/0/15 & 19/69/0/13 & 61/214/0/54 \\
\midrule
BGE-M3 & 1 & 0/115/0/28 & 0/72/0/13 & 0/82/0/19 & 0/269/0/60 \\
 & 5 & 11/125/0/7 & 10/73/0/2 & 10/86/0/5 & 31/284/0/14 \\
 & 20 & 16/122/0/5 & 16/68/0/1 & 14/86/0/1 & 46/276/0/7 \\
\midrule
ColBERTv2 & 1 & 0/127/0/16 & 0/82/0/3 & 0/94/0/7 & 0/303/0/26 \\
 & 5 & 13/126/0/4 & 13/72/0/0 & 13/87/0/1 & 39/285/0/5 \\
 & 20 & 25/116/0/2 & 21/64/0/0 & 20/81/0/0 & 66/261/0/2 \\
\midrule
Hybrid RRF & 1 & 0/97/1/45 & 0/67/0/18 & 0/80/0/21 & 0/244/1/84 \\
 & 5 & 12/113/0/18 & 11/64/0/10 & 11/81/0/9 & 34/258/0/37 \\
 & 20 & 20/116/0/7 & 20/61/0/4 & 18/81/0/2 & 58/258/0/13 \\
\midrule
RRF + rerank & 1 & 0/129/0/14 & 0/80/0/5 & 0/94/0/7 & 0/303/0/26 \\
 & 5 & 14/127/0/2 & 17/68/0/0 & 17/83/0/1 & 48/278/0/3 \\
 & 20 & 26/116/0/1 & 24/61/0/0 & 23/78/0/0 & 73/255/0/1 \\
\bottomrule
\end{tabular}
\caption{Complete object-distinct counts across every frozen search stack and
cutoff. Each cell lists both objects returned, selected object only, trace object
only, and neither object, in that order:
$P_1H_1/P_1H_0/P_0H_1/P_0H_0$. Model denominators are 143, 85, and 101.
The pooled denominator is 329 model-question records from 160 questions.
Pooled rows are descriptive because records sharing a question are dependent.}
\label{tab:mismatch-overlap}
\end{table*}

Across all 1,792 resolvable selections, the final reranker returns $U$ on 1,775
records and $G$ on 1,497. The labels differ on 278. Within the 329
object-distinct records, moving from body-only BM25 to the final reranker changes
74 selected-object labels from miss to hit and none from hit to miss. The
selected-object return rate rises by 22.5 points, with a 95\% question-cluster
bootstrap interval of [16.3, 29.3] points. This is a selected-name stress test.

The remaining tables expand the joint outcomes, omission mechanisms, and
complete cutoff grid.

\begin{table}[htbp]
\centering
\footnotesize
\setlength{\tabcolsep}{4.2pt}
\begin{tabular}{lrrr}
\toprule
Stack & $k=1$ & $k=5$ & $k=20$ \\
\midrule
Body BM25 & 209 & 213 & 214 \\
BGE-M3 & 269 & 284 & 276 \\
ColBERTv2 & 303 & 285 & 261 \\
Hybrid RRF & 245 & 258 & 258 \\
RRF + rerank & 303 & 278 & 255 \\
\bottomrule
\end{tabular}
\caption{Number of object-distinct records receiving different selected-object
and trace-object labels. Every cell has denominator 329.}
\label{tab:disagreement-grid}
\end{table}

\begin{table}[htbp]
\centering
\footnotesize
\setlength{\tabcolsep}{2.0pt}
\begin{tabular}{lrrrr}
\toprule
Model & Absent at 5 & Ranks 6 to 20 & Nonmatch & Unresolved \\
\midrule
Claude & 165 & 69 & 5 & 91 \\
GPT    & 151 & 59 & 5 & 87 \\
Gemini & 150 & 63 & 4 & 83 \\
\bottomrule
\end{tabular}
\caption{Partition of body-only BM25 top-five nonmembership. Absent at 5 is the
total, and the other three columns partition it. Nonmatch is verified when
fewer than 20 documents satisfy the OR query and the selected passage is
absent. A full 20-result list cannot separate a lower rank from nonmatch.}
\label{tab:bm25-omission-mechanisms}
\end{table}

\begin{table}[htbp]
\centering
\footnotesize
\setlength{\tabcolsep}{3.0pt}
\begin{tabular}{lrrrr}
\toprule
Retriever & $Q$ & Claude & GPT & Gemini \\
\midrule
BM25 & .183 & .580 & .648 & .637 \\
BGE-M3 & .183 & .745 & .827 & .812 \\
ColBERTv2 & .185 & .772 & .858 & .840 \\
Hybrid RRF & .230 & .702 & .782 & .767 \\
RRF + rerank & .358 & .777 & .868 & .852 \\
\bottomrule
\end{tabular}
\caption{Traced-passage $R@5$ across retrieval stacks. Every rate has
denominator 600. $Q$ uses the 600 unique original questions and is shared
across models. Each model column queries with that model's selected decoded-ID
query, with invalid upstream rows counted as misses.}
\label{tab:text-query-recall}
\end{table}

\runin{Results across every cutoff}
The full retrieval matrices show that query choice changes the returned IDs
beyond cutoff five. With BM25, a question query recalls the dataset trace at
.087/.278 for cutoffs 1/20. Queries built from the decoded selected ID give
.370 to .423 at cutoff 1 and .680 to .753 at cutoff 20. At cutoff five, question
recall is .183. The corresponding Claude/GPT/Gemini values are
.580/.648/.637 for decoded selected-ID queries, .590/.647/.637 for
model-written facts, and .667/.740/.645 for direct rewrites. Across BGE-M3,
ColBERTv2, hybrid RRF, and hybrid reranking, each model's decoded-ID query also
exceeds the question at both cutoffs. The replay retains the result for every
model, retriever, and cutoff.

The matched-five answer audit behind
Table~\ref{tab:delivery-switch-answer} uses complete terminal rows from the
frozen answer matrices. A record enters the comparison only when its
experimental arm, question and model pairing, five-passage context, no-hint prompt,
selected-passage return label, normalized exact-match score, denominator, and
question-cluster assignment match the frozen protocol.

\FloatBarrier

\subsection{Trace-deletion call settings}
\label{app:model-settings}

\begingroup
\raggedright
The trace-deletion experiment calls
\texttt{google/\allowbreak gemini-3.1-pro-preview} through an API gateway
pinned to Google AI Studio with provider fallback disabled. The settings are
temperature 1.0, medium reasoning, strict one-field JSON, no seed, three
repeats, and at most two attempts. Three successful preflight calls tested the
route and response schema and are excluded from analysis. The 561 main call
tasks were released together for parallel execution, and retries produced 570
provider requests; 24 tasks ended in failure. The intact arm sends the same
message content as the bound historical requests, using a different transport
and schema serialization.

The failures comprise 23 provider error envelopes and one invalid JSON response.
They are distributed as 13 intact, 8 trace-drop, and 3 sham-drop tasks. The
intact and sham F1 ordering changes under complete-case and valid-repeat
summaries. Both trace-removal effects remain.

\begin{table}[htbp]
\centering
\scriptsize
\setlength{\tabcolsep}{2.6pt}
\caption{Per-arm call accounting and repeat stability. Provider requests
include retries. Same answer means three valid identical normalized outputs,
and majority exact means an exact answer in at least two repetitions.}
\label{tab:deletion-accounting}
\begin{tabular}{@{}L{2.65cm}rrr@{}}
\toprule
& Intact & Trace-drop & Sham-drop \\
\midrule
Items & 64 & 64 & 59 \\
Scheduled calls & 192 & 192 & 177 \\
Provider requests & 194 & 194 & 182 \\
Valid / failed calls & 179 / 13 & 184 / 8 & 174 / 3 \\
Same valid answer, 3/3 & 51/64 & 43/64 & 54/59 \\
Majority exact & 63/64 & 9/64 & 57/59 \\
\bottomrule
\end{tabular}
\end{table}

The ranking manifest pins \texttt{BAAI/\allowbreak bge-m3} at revision
\texttt{5617a9f6}, \texttt{colbert-ir/\allowbreak colbertv2.0} at revision
\texttt{c1e84128}, and \texttt{BAAI/\allowbreak bge-reranker-v2-m3} at
revision \texttt{953dc6f6}. The corresponding model families are described in
\citep{chen2024bgem3,santhanam2022colbertv2,bge2024reranker}.

\par
\endgroup

\runin{Larger development retrieval check}
A separate 1,397-item cohort tests text projection across the same five stacks.
It contains 1,182 records without a model selection and 215 whose answer appears
in the table. Starting from the dataset trace, we compare original-question and
decoded-ID retrieval.

\begin{table}[htbp]
\centering
\footnotesize
\setlength{\tabcolsep}{3.5pt}
\begin{tabular}{lrrr}
\toprule
Retriever & $R@5(Q)$ & $R@5(\mathrm{ID})$ & Difference \\
\midrule
BM25 & .213 & .765 & .552 \\
BGE-M3 & .192 & .957 & .765 \\
ColBERTv2 & .236 & .992 & .756 \\
Hybrid RRF & .243 & .892 & .649 \\
Hybrid + rerank & .359 & .993 & .634 \\
\bottomrule
\end{tabular}
\caption{Cutoff-five trace return on the 1,397-item development cohort. $Q$
uses the original question, and ID uses the decoded dataset-traced identifier.}
\label{tab:broad-retrieval}
\end{table}

\subsection{Open-weight Qwen check}
\label{app:runtime}

The open-weight check asks whether the selected key survives the same retrieval
handoff with Qwen3-32B-AWQ upstream. We use $C=1$ when the selected object and
dataset trace resolve to the same object, and $C=0$ otherwise. Revision
\texttt{0499c3ac83fdef88} uses reasoning, a 4,096-token cap, temperature .6,
top-$p$ .95, top-$k$ 20, presence penalty 1.5, and global seed 12,345 with
deterministic per-task derivation. One parser-valid selected link falls outside
the displayed candidates and remains explicit.

\begin{table}[htbp]
\centering
\footnotesize
\setlength{\tabcolsep}{2.7pt}
\begin{tabular}{lrrrr}
\toprule
Subset & Resolvable & $k=1$ & $k=5$ & $k=20$ \\
\midrule
Final hybrid, all & 334 & 294 & 330 & 332 \\
Final hybrid, $C=1$ & 229 & 204 & 226 & 228 \\
Final hybrid, $C=0$ & 105 & 90 & 104 & 104 \\
\midrule
Body BM25, all & 334 & 165 & 231 & 282 \\
Body BM25, $C=1$ & 229 & 113 & 160 & 195 \\
Body BM25, $C=0$ & 105 & 52 & 71 & 87 \\
Exact decoded field & 334 & 334 & 334 & 334 \\
\bottomrule
\end{tabular}
\caption{Selected-object return in the open-weight check. Final hybrid uses
Qwen's recorded retrieval fact. Body BM25 uses the decoded selected ID and
reuses a hash-bound ranking only for an identical FTS5 query. The cutoff
columns report returned-object counts.}
\label{tab:qwen-exact-handoff}
\end{table}

\runin{Joint outcomes at cutoff five}
For all 334 resolvable records,
$(P_1H_1,P_1H_0,P_0H_1,P_0H_0)=(232,98,0,4)$. On the 105 $C=0$ records, the
same table is $(6,98,0,1)$. One additional $C=0$ key is unresolvable and has
$H=0$. Released hashes bind the rows, tasks, rankings, and resolver.

\end{document}